\documentclass{article} % For LaTeX2e

\usepackage{silence}
\usepackage[utf8]{inputenc} % From ACL template
\usepackage{times} % From ACL template
\usepackage{latexsym} % From ACL template
\usepackage[dvipsnames]{xcolor} % For color names
\usepackage{amsmath} % Extended math typesetting
\usepackage{amssymb} % Extended math symbols
\usepackage{mathtools} % More mathtools
\usepackage{annotate-equations} % Help annotate equations
\usepackage{pifont} % More symbols (also needed for checkmark and x-mark symbol macros)
\usepackage{truncate} % Help truncate text
\usepackage{enumitem} % allows setting leftmargin on \begin{enumerate} or \begin{itemize}
\usepackage{csquotes} % Inline and display quotes
\usepackage{xurl} % For \url
\usepackage[dont-mess-around]{fnpct} % Automatically ensures footnotes go after the punctuation mark
\usepackage{soul} % For text highlighting
\usepackage{ragged2e} % allows \centering
\usepackage{float} % allows the [H] on \begin{figure}[H]

\usepackage{caption}
\usepackage{subcaption} % For subfigures / captions: https://tex.stackexchange.com/a/122329

\usepackage{graphicx} % allows \includegraphics

\usepackage{array}
\usepackage{booktabs}
\usepackage{multirow}
\usepackage{tabularx}
\usepackage{tabulary}
\usepackage{makecell}

\usepackage{tikz,pgfplots}
\pgfplotsset{compat=1.17}

\usepackage{xinttools} % Useful to populate table data with a loop
\usepackage{ifthen} % If-then in LaTeX

\usepackage{lipsum}

\usepackage{todonotes}

\usepackage[T1]{fontenc}
\newcommand{\cmmnt}[1]{}

\usepackage{longtable}
\usepackage{xltabular}
\usepackage{setspace}

\newcommand{\fipipelinefigure}{
    \begin{figure*}[h!]
        \centering
        \includegraphics[width=\textwidth]{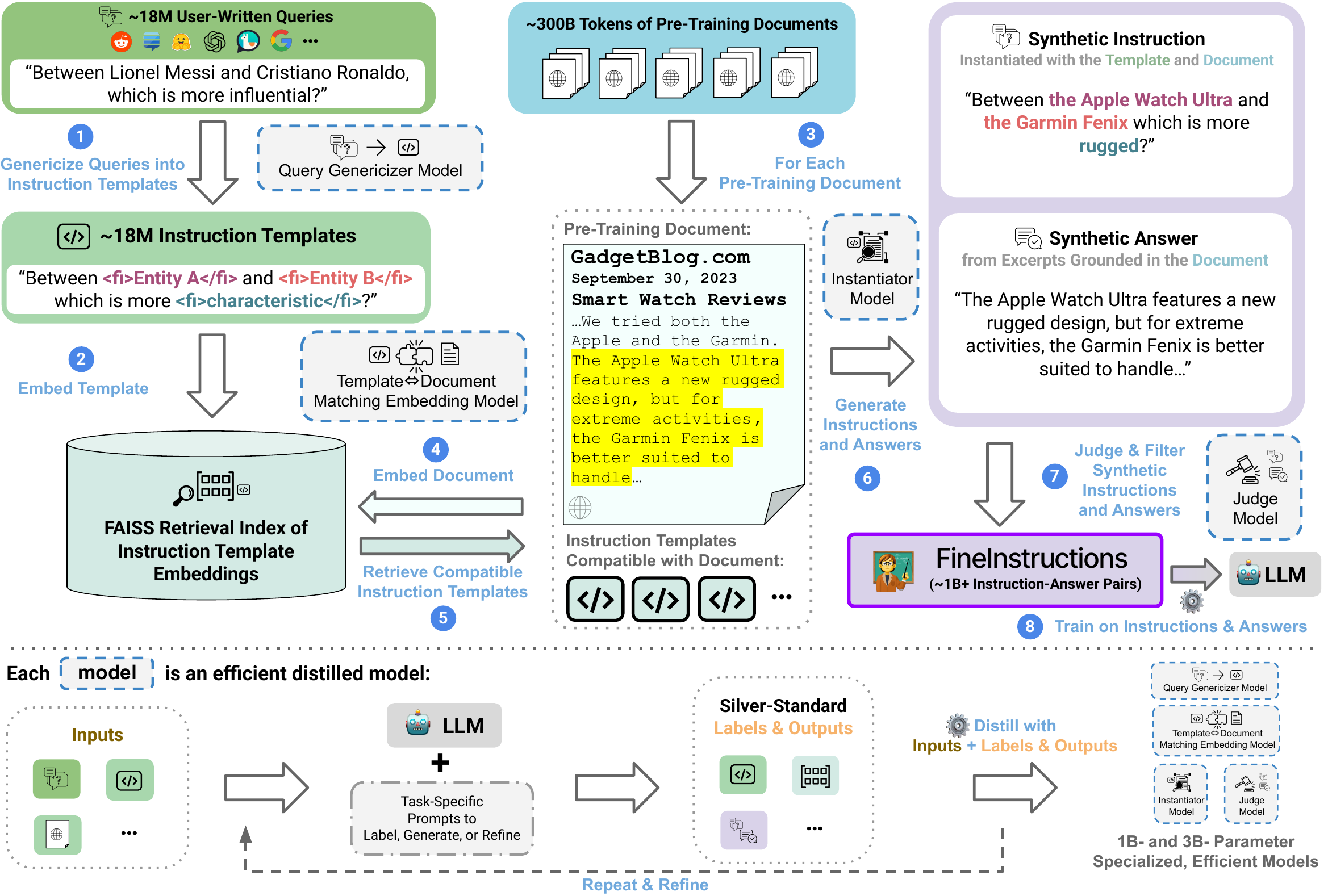}
        \caption{The FineInstructions pipeline for efficiently generating diverse, pre-training scale, synthetic instruction-answer pairs.}
        \label{fig:fipipeline}
    \end{figure*}
}

\newcommand{\verbdiversityfigure}{
    \begin{figure*}[h!]
        \centering
        \subfloat[FineInstructions (subset)]{
            \includegraphics[width=0.46\textwidth]{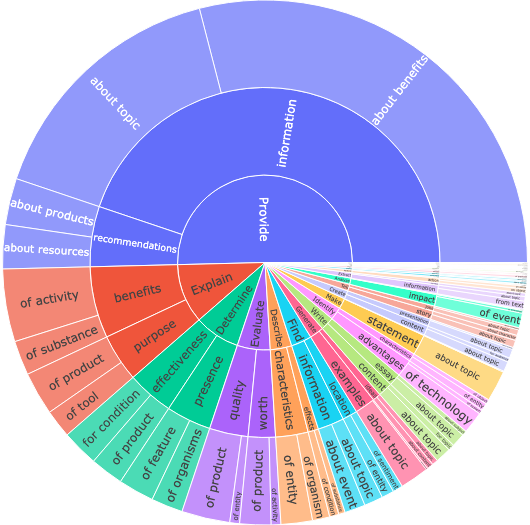}
            \label{fig:verbdiversity:a}
        }
        \hfill
        \subfloat[FineInstructions (medical subset)]{
            \includegraphics[width=0.46\textwidth]{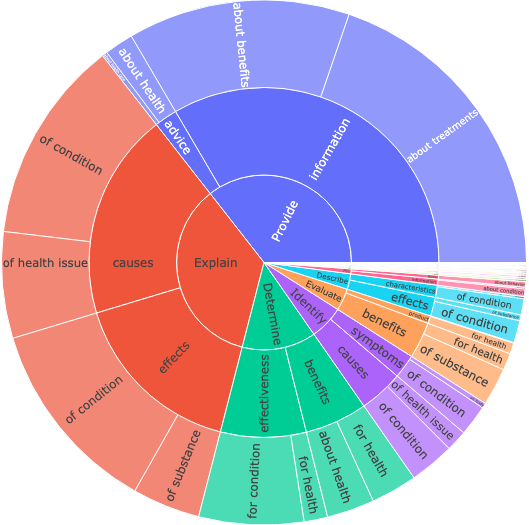}
            \label{fig:verbdiversity:b}
        }
        \caption{A visualization of the task diversity in a small subset of FineInstructions. We also preview domain-specific diversity, comparing with a subset that was instantiated from medicine-related templates. Charts can be read from the inner ring to the outer ring for each ``slice''.}
        \label{fig:verbdiversity}
    \end{figure*}
}

\newcommand{\fipromptexamplefigure}{
\begin{figure*}[h]
    \centering
    \fbox{%
        \begin{minipage}[t]{0.9\textwidth}
            \vspace{0.5em} % Top padding
            \scriptsize
Below is a generic query template that can be used for instruction-tuning an LLM. It contains a template of a natural language user query, with template variables using <fi></fi> tags (``<fi>description of content that should go there....</fi>'').
\\~\\\textit{\{\{example of a template...\}\}}
\\~\\These templates can then be instantiated from a *compatible* document (such as a web page, article, book, essay, etc.) that has contain enough information to both fill in all of the template variables and also has the answer to the query so that we can create grounded, realistic, diverse instruction-tuning questions and answers using a combination of the template + document.
\\~\\Given the provided template + a compatible document we found below:
\\~\\1. **Task:** Can you instantiate the template by filling in the <fi></fi> template variables to create a realistic instruction that could be realistically asked about the content/topic/subject described in the document?
\\~\\2. **Answerability and Template Incompatibility:** ...
\\~\\\textit{\{\{other requirements...\}\}}
\\~\\{<}{<}{<}Instruction Template:\\
\texttt{\{\{template\}\}}\\
{>}{>}{>}
\\~\\{<}{<}{<}Compatible Document:\\
\texttt{\{\{document\}\}}\\
{>}{>}{>}
            \vspace{0.5em} % Bottom padding
        \end{minipage}%
    }
    \label{fig:fipromptexample}
\end{figure*}
}

\newcommand{\customparbox}[3]{%
  \parbox[#1]{#2}{%
    \begin{spacing}{1.2} % adjust 1.2 to your desired line spacing
      \scriptsize
      #3
    \end{spacing}%
    \vspace{2em}
  }%
}

\newcommand{\categorydiversitytable}{
\begin{table}[h!]
\centering
\small
\begin{tabular}{lr}
\toprule
\textbf{Domain/Category} & \textbf{Percentage (\%)} \\
\midrule
Science & 36.61\% \\
Math & 0.58\% \\
Code & 0.25\% \\
Medicine & 10.40\% \\
Personal Life & 14.74\% \\ \hline
Reasoning Task & 10.99\% \\
Tasky vs. Q\&A & 6.42\% \\
\bottomrule
\end{tabular}
\caption{1B+ instructions from FineInstructions classified into various categories.}
\label{table:categorydiversity}
\end{table}
}

\newcommand{\fibenchmarktablescores}{
\begin{table*}[h!]
\centering
\footnotesize
\setlength{\tabcolsep}{4pt}
\begin{tabular}{p{4cm}cccc}
\toprule
\multirow{2}{*}{\textbf{Method}} & 
\multicolumn{2}{c}{\textbf{MixEval}} & 
\textbf{MT-Bench-101} & 
\textbf{AlpacaEval} \\
& \multicolumn{2}{c}{\textit{Acc (\%)}} & \textit{Likert Score} & \textit{FI Win Rate \% ($\Delta$ Win Margin \%)} \\
\cmidrule(lr){2-3}
& \textbf{Standard} & \textbf{Hard} & & \\
\midrule
\multicolumn{2}{l}{\textit{IPT Corpus (23B)}} & & & \\ \midrule
Standard Pre-Training & 17.8 & 14.0 & 1.9 & \textbf{73.6\%} ($\Delta$ \textbf{47.2}\%) \\
IPT & 19.8 & 16.7 & 2.4 & \textbf{68.2\%} ($\Delta$ \textbf{36.4}\%) \\
FineInstructions & \textbf{31.7} & \textbf{19.2} & \textbf{2.8} & -- \\
\midrule
\multicolumn{2}{l}{\textit{Nemotron-CC Corpus (300B)}} & & & \\ \midrule
Standard Pre-Training & 24.0 & 17.1 & 3.5 &  \textbf{63.6\%} ($\Delta$ \textbf{27.2}\%) \\
WRAP & 22.8 & 18.4 & 3.6 & \textbf{65.1\%} ($\Delta$ \textbf{30.2}\%) \\
Q\&A & 27.1 & 18.9 & 3.4 & \textbf{76.1\%} ($\Delta$ \textbf{52.2}\%) \\
Nemotron-CC & 24.5 & 16.7 & 3.6 & \textbf{65.9\%} ($\Delta$ \textbf{31.8}\%) \\
FineInstructions & \textbf{33.0} & \textbf{21.8} & \textbf{3.9} & -- \\ 
\bottomrule
\end{tabular}
\caption{Benchmark performance of 1.8B parameter models pre-trained on FineInstructions and various baselines. We \textbf{bold} the strongest method and \textbf{bold} the win rate \%  if FineInstructions wins.}
\label{table:fibenchmarktablescores}
\end{table*}
}

\newcommand{\fibenchmarktablejudgeablation}{
\begin{table*}[h!]
\centering
\footnotesize
\setlength{\tabcolsep}{4pt}
\begin{tabular}{p{3.6cm}ccccc}
\toprule
\multirow{2}{*}{\textbf{Method}} & 
\multicolumn{2}{c}{\textbf{MixEval}} & 
\textbf{MT-Bench-101} & 
\multicolumn{2}{c}{\textbf{AlpacaEval}} \\
& \multicolumn{2}{c}{\textit{Acc (\%)}} & \textit{Likert Score} & \multicolumn{2}{c}{\textit{FI Win Rate \% ($\Delta$ Win Margin \%)}}  \\
\cmidrule(lr){2-3} \cmidrule(lr){5-6}
& \textbf{Standard} & \textbf{Hard} & & \textbf{Without Judging} & \textbf{With Judging} \\
\midrule
\multicolumn{2}{l}{\textit{IPT Corpus (23B)}} & & & \\ \midrule
Standard Pre-Training & 17.8 & 14.0 & 1.9 & 72.1\% ($\Delta$ 44.2\%) & \textbf{73.6\%} ($\Delta$ \textbf{47.2}\%) \\
IPT & 19.8 & 16.7 & 2.4 & 61.7\% ($\Delta$ 23.4\%) & \textbf{68.2\%} ($\Delta$ \textbf{36.4}\%) \\
FineInstructions & 30.1 & \textbf{20.2} & 2.6 & -- & 46.5\% ($\Delta$ -7\%) \\
FineInstructions (judged) & \textbf{31.7} & 19.2 & \textbf{2.8} & -- & -- \\
\midrule
\multicolumn{2}{l}{\textit{Nemotron-CC Corpus (300B)}} & & & \\ \midrule
Standard Pre-Training & 24.0 & 17.1 & 3.5 &  54.9\% ($\Delta$ 9.8\%) & \textbf{63.6\%} ($\Delta$ \textbf{27.2}\%) \\
WRAP & 22.8 & 18.4 & 3.6 & 58.4\% ($\Delta$ 16.8\%) & \textbf{65.1\%} ($\Delta$ \textbf{30.2}\%) \\
Q\&A & 27.1 & 18.9 & 3.4 & 68.7\% ($\Delta$ 37.4\%) & \textbf{76.1\%} ($\Delta$ \textbf{52.2}\%) \\
Nemotron-CC & 24.5 & 16.7 & 3.6 & 55.7\% ($\Delta$ 11.4\%) & \textbf{65.9\%} ($\Delta$ \textbf{31.8}\%) \\
FineInstructions & \textbf{33.3} & \textbf{21.8} & 3.6 & -- & \textbf{61.4\%} ($\Delta$ \textbf{22.8}\%) \\
FineInstructions (judged) & 33.0 & \textbf{21.8} & \textbf{3.9} & -- & -- \\
\bottomrule
\end{tabular}
\caption{Incorporating a judging and filtering stage on top of FineInstructions leads to further improvements in performance.}
\label{table:fibenchmarktablejudgeablation}
\end{table*}
}

\newcommand{\modelsizeexperimentscores}{
\begin{table*}[h!]
\centering
\setlength{\tabcolsep}{4pt}
\scriptsize
\begin{tabular}{p{4cm}ccccc}
\toprule
\multirow{2}{*}{\textbf{Method}} & 
\multicolumn{2}{c}{\textbf{MixEval}} & 
\textbf{MT-Bench-101} & 
\multicolumn{2}{c}{\textbf{AlpacaEval}} \\
& \multicolumn{2}{c}{\textit{Acc (\%)}} & \textit{Likert Score} & \multicolumn{2}{c}{\textit{FI Win Rate \% ($\Delta$ Win Margin \%)}} \\
\cmidrule(lr){2-3}
\cmidrule(lr){5-6}
& \textbf{Standard} & \textbf{Hard} & & \textbf{One Model Size Down} & \textbf{Same Model Size}\\
\midrule
\multicolumn{2}{l}{\textit{Nemotron-CC Corpus (300B) -- 300M Parameter Models}} & & & \\ \midrule
Standard Pre-Training & 18.5 & 14.3 & 2.8 & -- & \textbf{65.1\%} ($\Delta$ \textbf{30.2}\%) \\
Nemotron-CC & 18.7 & 13.9 & 3.0 & -- & \textbf{57.5\%} ($\Delta$ \textbf{15.0}\%) \\
FineInstructions & \textbf{31.9} & \textbf{22.0} & \textbf{3.2} & -- & -- \\ \midrule
\multicolumn{2}{l}{\textit{Nemotron-CC Corpus (300B) -- 1.8B Parameter Models}} & & & \\ \midrule
Standard Pre-Training & 24.0 & 17.1 & 3.5 &  49.2\% ($\Delta$  -1.6\%) & \textbf{63.6\%} ($\Delta$ \textbf{27.2}\%) \\
Nemotron-CC & 24.5 & 16.7 & 3.6 & 49.8\% ($\Delta$  -0.4\%) & \textbf{65.9\%} ($\Delta$ \textbf{31.8}\%) \\
FineInstructions & \textbf{33.0} & \textbf{21.8} & \textbf{3.9} & -- & -- \\ \midrule
\multicolumn{2}{l}{\textit{Nemotron-CC Corpus (300B) -- 7B Parameter Models}} & & & \\ \midrule
Standard Pre-Training & 33.0 & 21.6 & \textbf{4.4} &  50.0\% ($\Delta$  0.0\%)  & \textbf{59.0\%} ($\Delta$ \textbf{18.0}\%) \\
Nemotron-CC & 32.7 & 21.5 & 4.3 &  45.6\% ($\Delta$  -8.8\%)  & \textbf{56.6\%} ($\Delta$ \textbf{13.2}\%) \\
FineInstructions & \textbf{42.6} & \textbf{25.8} & 4.3 & -- & -- \\
\bottomrule
\end{tabular}
\caption{Pre-training results for models with 300M, 1.8B, and 7B parameters. At a fixed model size, models pre-trained on FineInstructions are competitive with or outperform, baselines trained at the next larger scale (e.g., 300M $\rightarrow$ 1.8B, 1.8B $\rightarrow$ 7B). Under equivalent token and compute budgets, FineInstructions consistently achieves higher performance across MixEval, MT-Bench-101, and AlpacaEval, indicating that it is an effective data mixture for training smaller, more compute-efficient models with stronger capabilities.}
\label{table:modelsizeexperimentscores}
\end{table*}
}

\newcommand{\fiillustratedexamplesfigure}{
\newcolumntype{P}[1]{>{\RaggedRight\arraybackslash}p{##1}}
\begin{small} 
\begin{xltabular}{.9\textwidth}{P{0.3\textwidth} P{0.3\textwidth} P{0.3\textwidth}}
\endfirsthead

\multicolumn{3}{c}{} \\ % No header
\endhead

 \multicolumn{3}{c}{{\textit{Continued on next page...}}} \\
\endfoot

\endlastfoot

\toprule
\multicolumn{3}{p{0.9\textwidth}}{\textbf{Vanilla Pre-Training Document:}} \\
\multicolumn{3}{p{0.9\textwidth}}{\customparbox{t}{0.95\textwidth}{Structure and Synchronicity for Better Charting\\
Two key characteristics will help you to ensure that your notes communicate not only what you did, but also what you were thinking.\\
Fam Pract Manag. 2011 Jul-Aug;18(4):15-17.\\~\\
How many times have you read a medical note that does not make the selection of the diagnosis or treatment clear? Have you ever read your own notes after receiving notice of a malpractice suit and winced at the inconsistencies? Poorly constructed medical notes are a widespread problem. I've seen it while reviewing the charts of medical students and residents to ensure that they met the standard of care and avoided malpractice risk. While most physicians document history and physical data with the required number of CPT elements, few clearly convey a line of reasoning that reveals their clinical thought process. A note that documents a detailed history and moderately complex decision making does not necessarily illuminate why certain decisions were reached or why a particular treatment was justified.\\~\\
I wanted to find out where in the educational process students learned key documentation concepts, so I did some research that included informally querying medical students from multiple medical schools who were rotating at our residency. None of them could describe learning a formal structure for completing assessments and plans, which is consistent with a study that found only 4 percent of standardized encounters were accurately charted by medical students.1 I also discovered that the 2010 United States Medical Licensing Examination clinical skills exam guide states that students are expected to present a list of differential diagnoses in order of likelihood along with desired evaluations, but no requirements existed for discussing the clinical rationale. As a result of these and other findings, I developed a formal framework, described in this article, to teach residents and students an appropriate way to construct their notes. The initial feedback has been highly positive.\\~\\
Documenting a confirmed diagnosis\\~\\
To begin, let's review what should be included when documenting a confirmed diagnosis. Generally, six elements are needed...}} \\ \midrule
\textbf{Instruction Template} & \textbf{Synthetic Instruction} & \textbf{Synthetic Answer} \\ \midrule

\customparbox{t}{0.3\textwidth}{I'm looking for a template or guidelines for how to write a \texttt{<fi>type of document, e.g. technical report, research paper, etc.</fi>}.

The \texttt{<fi>type of document</fi>} is \texttt{<fi>context or field of study</fi>}-related and intended for a \texttt{<fi>role or position</fi>} to review the work of the \texttt{<fi>task or investigation</fi>} of a \texttt{<fi>type of issue or problem</fi>} related to \texttt{<fi>field or industry</fi>}.} & \customparbox{t}{0.3\textwidth}{I'm looking for a template or guidelines for how to write a medical note.\\~\\The medical note is healthcare-related and intended for a physician to review the work of the clinical investigation of a medical issue related to patient care.} & \customparbox{t}{0.3\textwidth}{To write a medical note, consider the following guidelines. To begin, let's review what should be included when documenting a confirmed diagnosis. Generally, six elements are needed...}  \\

\toprule
\multicolumn{3}{p{0.9\textwidth}}{\textbf{Vanilla Pre-Training Document:}} \\
\multicolumn{3}{p{0.9\textwidth}}{\customparbox{t}{0.95\textwidth}{Spot writes: <<<<your dentist is correct diabetes is often accomnpied by severe periodontal disease and complicated by xerosomia (xerostomia, dry mouth)this results in increased incidence of caries and loss of teeth. also infections>>>>> Jack writes: <<<<>>>> Spot and Jack, I know all of this! No, Jack she didn't see the start of periodontal disease, had the full workup and absolutely no sign of it at all. I had one cavity in a flossing area which prompted the extra flossing advice. My whole point was that with a busy life + DM, extra flossing was going to be low on my list of extras to accomplish. I already floss daily, just not 3-4 times a day and after every meal. BTW, my flossing time is way more than 2-3 minutes Jack. Sooooo, I understand the DANGERS of not flossing more often than nonD folks, and if all of my teeth fall out in 50 years when I'm 80, Jack and Spot can say, ``Told you so.'' (vbg) Jill (Who Is Still Not Flossing 5 times a Day)\\~\\
------------------------------------------------ Do You Yahoo!? Yahoo! Auctions - Buy the things you want at great prices. ------------------------------------------------ for HELP or to subscribe/unsubscribe, contact: HELP@insulin-pumpers.org send a DONATION}} \\ \midrule
\textbf{Instruction Template} & \textbf{Synthetic Instruction} & \textbf{Synthetic Answer} \\ \midrule
\customparbox{t}{0.3\textwidth}{what effects does \texttt{<fi>a medical or psychological condition</fi>} have?} & \customparbox{t}{0.3\textwidth}{what effects does diabetes have?} & \customparbox{t}{0.3\textwidth}{Diabetes has several effects, including being often accompined by severe periodontal disease and complicated by xerosomia (xerostomia, dry mouth)this results in increased incidence of caries and loss of teeth.}  \\

\toprule
\multicolumn{3}{p{0.9\textwidth}}{\textbf{Vanilla Pre-Training Document:}} \\
\multicolumn{3}{p{0.9\textwidth}}{\customparbox{t}{0.95\textwidth}{

This article is the first part of building a production grade nodejs,graphql and typescript server with auto deployment pipeline.
there are tons of articles and tutorials are available to explain how to build a scalable backend solutions. but, what i felt missing was a solution that connects all the dots with a simple and proper use-cases.
Most of the time, we won't directly jump into building the application by coding it. we would start with designing how the backend should be based on the application requirement. knowing how to do that is an important skill to have than just knowing the syntax of x or y framework.\\~\\
Before coding, let's see what we are going to build to learn the concepts. we will be building a Image management application where users can manage their image resources.
Mainly, it will contain login,signup and main application. In main application, user can upload images and view all the uploaded images. Even though it's a simple use-case, we can understand all the concepts using this simple use-case.\\~\\
Login Screen\\
Main App Screen\\~\\
Now, we know what application we are going to build. let's apply all the best practices to build a Nodejs,GraphQL and TypeScript Server.\\~\\
This part is going to explain how to setup the application and create a query and mutations using GraphQL and TypeScript.
we will be using type-graphql for your application. let's install the required dependancies for our application, npm i apollo-server-express express graphql reflect-metadata type-graphql mongoose....

}} \\ \midrule
\textbf{Instruction Template} & \textbf{Synthetic Instruction} & \textbf{Synthetic Answer} \\ \midrule
\customparbox{t}{0.3\textwidth}{how to \texttt{<fi>a few word description of the action to be performed</fi>} \texttt{<fi>name of a service, system, or platform</fi>}?} & \customparbox{t}{0.3\textwidth}{how to create a GraphQL schema and connect to a MongoDB database using type-graphql?} & \customparbox{t}{0.3\textwidth}{To create a GraphQL schema and connect to a MongoDB database using type-graphql, follow these steps:\\
we will be using type-graphql for your application. let's install the required dependancies for our application,
1npm i apollo-server-express express graphql reflect-metadata type-graphql mongoose...}  \\

\bottomrule

\end{xltabular}
\end{small}

}

\newcommand{\mixevalsubbenchmarkstable}{
\begin{table}[h!]
\centering
\footnotesize
\setlength{\tabcolsep}{5pt}
\begin{tabular}{l r ccc ccccc}
\toprule
& & \multicolumn{3}{c}{\textbf{IPT Corpus (23B)}} & \multicolumn{5}{c}{\textbf{Nemotron-CC Corpus (300B)}} \\
\cmidrule(lr){3-5} \cmidrule(lr){6-10}
\textbf{Sub-Benchmark} & \textbf{$n$} & \textbf{Std. PT} & \textbf{IPT} & \textbf{FI} & \textbf{Std. PT} & \textbf{WRAP} & \textbf{Q\&A} & \textbf{Nemo-CC} & \textbf{FI} \\
\midrule
Overall & & 17.8 & 19.8 & \textbf{31.7} & 24.0 & 22.8 & 27.1 & 24.5 & \textbf{33.0} \\
\midrule
TriviaQA & 1328 & 3.2 & 6.7 & \textbf{13.2} & 20.4 & 16.8 & \textbf{23.0} & 20.3 & 20.9 \\
MMLU & 681 & 30.1 & 28.0 & \textbf{47.0} & 28.0 & 29.7 & 33.5 & 27.8 & \textbf{45.2} \\
DROP & 473 & 5.2 & 11.9 & \textbf{33.2} & 13.6 & 14.0 & 19.9 & 13.7 & \textbf{38.0} \\
HellaSwag & 308 & 32.8 & 29.2 & \textbf{43.5} & 23.7 & 28.6 & 28.9 & 36.7 & \textbf{37.7} \\
CommonsenseQA & 202 & 30.7 & 26.2 & \textbf{48.0} & 22.8 & 22.3 & 29.7 & 24.3 & \textbf{39.1} \\
MMLU-Pro & 185 & 15.1 & 16.8 & \textbf{33.5} & 15.7 & 10.8 & 14.1 & 16.2 & \textbf{25.9} \\
BoolQ & 171 & 53.8 & 55.0 & \textbf{55.6} & 56.1 & 50.9 & 45.6 & 48.0 & \textbf{59.1} \\
AGIEval & 121 & 14.8 & 25.9 & \textbf{28.7} & 25.0 & \textbf{27.8} & 26.9 & 20.4 & 23.1 \\
BBH & 115 & 10.3 & \textbf{40.2} & 22.8 & \textbf{25.0} & 15.3 & 16.8 & \textbf{25.0} & 24.3 \\
PIQA & 105 & 41.0 & 36.2 & \textbf{56.2} & 52.4 & 54.3 & 56.2 & 50.5 & \textbf{61.9} \\
SIQA & 93 & 33.3 & 32.3 & \textbf{38.7} & \textbf{41.9} & 35.5 & 37.6 & 39.8 & 31.2 \\
ARC & 91 & 29.7 & 30.8 & \textbf{50.5} & 20.9 & 30.8 & 40.7 & 26.4 & \textbf{49.5} \\
OpenBookQA & 43 & \textbf{51.2} & 25.6 & 44.2 & 37.2 & 25.6 & 32.6 & 27.9 & \textbf{48.8} \\
GSM8k & 40 & 0.0 & \textbf{1.3} & 0.0 & 0.0 & 0.0 & \textbf{2.8} & 0.5 & 1.3 \\
MATH & 31 & 0.0 & 0.0 & 0.0 & 0.0 & 0.0 & \textbf{3.2} & 0.0 & 0.0 \\
GPQA & 4 & \textbf{25.0} & 0.0 & 0.0 & 25.0 & 0.0 & 50.0 & 0.0 & \textbf{75.0} \\
WinoGrande & 4 & 0.0 & 25.0 & \textbf{75.0} & 50.0 & 25.0 & 25.0 & \textbf{75.0} & 0.0 \\
MBPP & 3 & 66.7 & 66.7 & \textbf{100.0} & 66.7 & 33.3 & 66.7 & 33.3 & \textbf{100.0} \\
HumanEval & 2 & \textbf{100.0} & \textbf{100.0} & \textbf{100.0} & 50.0 & 50.0 & 50.0 & \textbf{100.0} & \textbf{100.0} \\
\bottomrule
\end{tabular}
\caption{Detailed breakdown of MixEval (Standard) sub-benchmark performance (\% accuracy), with per-sub-benchmark sample counts ($n$). Methods are abbreviated: Std. PT (Standard Pre-Training), FI (FineInstructions), Nemo-CC (Nemotron-CC). We \textbf{bold} the strongest method within each corpus group.}
\label{table:mixevalsubbenchmarks}
\end{table}
}

\newcommand{\mixevalhardsubbenchmarkstable}{
\begin{table}[h!]
\centering
\footnotesize
\setlength{\tabcolsep}{5pt}
\begin{tabular}{l r ccc ccccc}
\toprule
& & \multicolumn{3}{c}{\textbf{IPT Corpus (23B)}} & \multicolumn{5}{c}{\textbf{Nemotron-CC Corpus (300B)}} \\
\cmidrule(lr){3-5} \cmidrule(lr){6-10}
\textbf{Sub-Benchmark} & \textbf{$n$} & \textbf{Std. PT} & \textbf{IPT} & \textbf{FI} & \textbf{Std. PT} & \textbf{WRAP} & \textbf{Q\&A} & \textbf{Nemo-CC} & \textbf{FI} \\
\midrule
Overall & & 14.0 & 16.7 & \textbf{19.2} & 17.1 & 18.4 & 18.9 & 16.7 & \textbf{21.8} \\
\midrule
TriviaQA & 331 & 1.9 & 5.8 & \textbf{5.9} & 9.4 & 9.4 & \textbf{11.8} & 11.0 & 9.8 \\
MMLU & 152 & 27.0 & 27.0 & \textbf{27.6} & 23.7 & 30.9 & 26.3 & 27.0 & \textbf{33.6} \\
DROP & 122 & 3.6 & 12.1 & \textbf{29.2} & 12.3 & 8.3 & 16.6 & 6.6 & \textbf{35.3} \\
HellaSwag & 61 & \textbf{29.5} & 26.2 & 24.6 & 26.2 & \textbf{27.9} & \textbf{27.9} & 18.0 & 24.6 \\
CommonsenseQA & 51 & 23.5 & 27.5 & \textbf{29.4} & 15.7 & 19.6 & 23.5 & \textbf{35.3} & 27.5 \\
MMLU-Pro & 90 & 11.1 & 8.9 & \textbf{15.6} & 11.1 & 11.1 & 11.1 & 10.0 & \textbf{17.8} \\
BoolQ & 32 & \textbf{53.1} & 46.9 & 43.8 & 50.0 & 53.1 & 53.1 & \textbf{59.4} & 46.9 \\
AGIEval & 65 & 15.8 & \textbf{24.6} & \textbf{24.6} & 21.1 & \textbf{24.6} & 19.3 & 12.3 & 19.3 \\
BBH & 20 & 5.0 & \textbf{31.5} & 15.0 & \textbf{18.0} & 12.5 & 12.5 & 14.0 & 9.0 \\
PIQA & 12 & \textbf{58.3} & 50.0 & 25.0 & 58.3 & 50.0 & \textbf{75.0} & 41.7 & 33.3 \\
SIQA & 26 & \textbf{38.5} & 30.8 & 34.6 & 26.9 & \textbf{38.5} & 23.1 & 26.9 & 26.9 \\
ARC & 6 & 0.0 & \textbf{16.7} & \textbf{16.7} & \textbf{50.0} & \textbf{50.0} & \textbf{50.0} & 16.7 & \textbf{50.0} \\
OpenBookQA & 8 & 25.0 & 12.5 & \textbf{37.5} & 37.5 & \textbf{62.5} & 12.5 & 25.0 & 37.5 \\
GSM8k & 7 & 0.0 & \textbf{14.3} & \textbf{14.3} & 0.0 & 0.0 & 0.0 & 0.0 & \textbf{14.3} \\
MATH & 12 & 0.0 & 0.0 & 0.0 & 0.0 & 0.0 & 0.0 & 0.0 & 0.0 \\
GPQA & 3 & 0.0 & 33.3 & \textbf{66.7} & \textbf{33.3} & 0.0 & 0.0 & 0.0 & 0.0 \\
WinoGrande & 2 & \textbf{100.0} & 0.0 & 50.0 & \textbf{100.0} & 50.0 & 50.0 & 0.0 & 0.0 \\
MBPP & --- & --- & --- & --- & --- & --- & --- & --- & --- \\
HumanEval & --- & --- & --- & --- & --- & --- & --- & --- & --- \\
\bottomrule
\end{tabular}
\caption{Detailed breakdown of MixEval-Hard sub-benchmark performance (\% accuracy), with per-sub-benchmark sample counts ($n$). Methods are abbreviated: Std. PT (Standard Pre-Training), FI (FineInstructions), Nemo-CC (Nemotron-CC). We \textbf{bold} the strongest method within each corpus group.}
\label{table:mixevalhardsubbenchmarks}
\end{table}
}

\usepackage[final]{colm2026_conference}

\usepackage{microtype}
\usepackage{hyperref}
\usepackage{url}
\usepackage{booktabs}

\definecolor{darkblue}{rgb}{0, 0, 0.5}
\hypersetup{colorlinks=true, citecolor=darkblue, linkcolor=darkblue, urlcolor=darkblue}

\title{FineInstructions: Scaling Synthetic Instructions to \\Pre-Training Scale}

\author{
Ajay Patel\thanks{Corresponding author: \texttt{patel.ajay285@gmail.com}} \\
University of Pennsylvania \\
\And
Colin Raffel \\
University of Toronto \\
Vector Institute \\
Hugging Face \\
\And
Chris Callison-Burch \\
University of Pennsylvania \\
}

\begin{document}

\maketitle

% Main sections
\begin{abstract}
Due to limited supervised training data, large language models (LLMs) are typically pre-trained via a self-supervised ``predict the next word'' objective on a vast amount of unstructured text data. To make the resulting model useful to users, it is further trained on a far smaller amount of ``instruction-tuning'' data comprised of supervised training examples of instructions and responses. To overcome the limited amount of supervised data, we propose a procedure that can transform the knowledge in internet-scale pre-training documents into billions of synthetic instruction and answer training pairs. The resulting dataset, called FineInstructions, uses \textasciitilde 18M instruction templates created from real user-written queries and prompts. These instruction templates are matched to and instantiated with human-written source documents from unstructured pre-training corpora. With ``supervised'' synthetic training data generated at this scale, an LLM can be pre-trained from scratch solely with the instruction-tuning objective, which is far more in-distribution with the expected downstream usage of LLMs (responding to user prompts). We conduct controlled token-for-token training experiments and find pre-training on FineInstructions outperforms standard pre-training and other proposed synthetic pre-training techniques on standard benchmarks measuring free-form response quality. Our resources can be found at \url{https://huggingface.co/fineinstructions}.
\end{abstract}

\section{Introduction}
\label{sec:introduction}

During self-supervised pre-training, LLMs are trained using a language modeling task like next token prediction over a large amount of text data.
This stage of training is where models acquire the vast majority of their knowledge and where the majority of compute, resources, and time is spent \citep{t5,gpt3,kaplan2020scalinglawsneurallanguage}. A pre-trained LLM can then be adapted to have better instruction-following capabilities by being further trained on a relatively small amount of supervised instruction-answer examples in a process known as instruction-tuning \citep{instructgpt,flan,t0,naturalinstructions}.
Existing instruction-tuning datasets have various issues.
Many are relatively small, consisting of a few thousand examples \citep{dolly,source_HuggingFaceH4_no_robots}.
Others are narrow and unrealistic, consisting of academic NLP tasks converted into instruction-tuning formats with a relatively small number of task templates \citep{t0,flan,naturalinstructions}.
Frontier language models have been used to generate large quantities of more diverse instruction-answer examples, but this has been shown to ultimately only help mimic those models superficially through distillation \citep{alpaca,orca,Honovich:22unnatural,motivation_imitationlimits}. These issues limit the instruction-tuning stage of LLMs, making it primarily useful for helping the model learn to follow instructions and learn response styles.
Consequently, the self-supervised pre-training stage is responsible for encoding the vast majority of knowledge in the model weights \citep{motivation_lima,instructiontuninglimits1,motivation_responsetuning}. %Instruction-tuning, however, makes a model vastly more useful for downstream use cases as it involves supervised learning on a dataset format in-distribution with downstream usage by users (a response conditioned on a query or instruction) instead of self-supervised learning on a language modeling objective.

% \mainfig

\fipipelinefigure

Beyond encoding knowledge, pre-training corpora have been shown to help models perform tasks via indirect supervision from examples of tasks that appear in the pre-training documents \citep{chen-etal-2024-parallel}. It is unclear, however, whether predicting the next token over pre-training documents is the most optimal or efficient way for models to absorb such capabilities. Recently proposed synthetic rephrasing and transformation pipelines \citep{wrap,nemotron} demonstrate transforming the original pre-training documents into alternative formats can improve both efficiency of knowledge absorption during pre-training and the performance and capabilities of the trained models.

In this paper, we introduce a procedure called FineInstructions (illustrated in Figure \ref{fig:fipipeline}) that transforms pre-training corpora into large collections of diverse and realistic instruction-response pairs.
Our pipeline pairs documents with instructions or queries a user might ask about the knowledge in the document and then extracts an answer or response grounded in the document. Instructions are created by templatizing \textasciitilde 18M queries written by real users which are then instantiated on a per-document basis. Converting pre-training documents into these synthetic instruction and answer pairs allows us to effectively perform supervised instruction-tuning at pre-training scale. We hypothesize such a restructuring of the data may aid knowledge absorption and model performance as prior studies have shown multi-task datasets and instruction-answer formats induce generalization and instruction-following capabilities \citep{t5,instructgpt,motivation_responsetuning}. This kind of restructuring also prevents pre-training compute budget from being wasted on fitting to potentially low-quality noise in documents (footers, headers, less educational content, etc.) and better utilized towards higher-quality, more educational sections of a document. Finally, it brings pre-training data further in-distribution with the expected downstream usage by users. Ultimately, we demonstrate that the synthetic instruction-answer pairs generated by our approach significantly differ from prior approaches in diversity, complexity, and quality. In sum, our contributions are:
\begin{enumerate}
    \item We created a large-scale dataset of \textasciitilde18M instruction templates. The templates were created by mining real user-written queries and tasks and converting them into generic templates. This large set of templates is what allows us to generate highly diverse and in-distribution synthetic data representative of real user tasks.
    \item We introduce the FineInstructions procedure for converting  pre-training documents into synthetic instructions and answers at scale. %The procedure involves matching documents with compatible instruction templates and generating synthetic instructions and queries in-distribution with the kind of queries real users may ask by instantiating instruction-templates and producing grounded responses by extracting an answer from the document.
    \item We demonstrate pre-training solely on FineInstructions outperforms standard pre-training and other proposed synthetic transformation pipelines on three LLM evaluation benchmarks correlated with human judgements of model response quality that span academic tasks and more realistic user tasks.
    \item We release our code, trained models, and FineInstructions, a dataset of 1B+ synthetic instruction and answer pairs, useful for training LLMs.
\end{enumerate}

\section{Related Work}

While LLMs are primarily trained on unstructured text sourced from web pages, books, and other natural sources, a number previous works have explored incorporating synthetic training data generated by an LLM itself as part of the pre-training process \citep{gunasekar2023textbooks,benallal2024cosmopedia,phi4,kimi,entigraph}. Other approaches have looked at filtering documents \citep{refinedweb,fineweb,dclm,nemotron} with LLM judgments of quality or re-weighting and re-sampling the mixture of documents  \citep{yadlowsky2023pretraining,efficientonlinedatamixing,doremi,dclm,mixinglaws,weborganizer}. Recent approaches have applied more sophisticated techniques, converting raw pre-training documents into instruction-answer pairs and other task formats \citep{ipt,nemotron,restructuredpretraining} or rephrasing low-quality documents into a higher-quality, clean documents \citep{wrap, nemotron,recyclingtheweb}. In our work, we generate synthetic training data by automatically creating and instantiating instruction templates. Other work has explored synthetically generating instruction-tuning data and reasoning data with LLMs and prompting \citep{Honovich:22unnatural,selfinstruct,alpaca,instructionbacktranslation,glan,lambert2024tulu,orca,openorca,selfconsistencycot,o1,deepseek}. \citet{Köksal:24longform} generate instruction-answer pairs for long-form generation, pairing pre-training documents with synthetic instructions that would plausibly generate them. Using instruction templates is also a common approach to creating instruction-tuning datasets \citep{bach2022promptsource,t0,flan,naturalinstructions,supernaturalinstructions,cookbook}. Approaches that simply prompt LLMs to transform or rephrase documents without templates can yield challenges with diversity in the synthetic data \citep{ge2024scaling,attrprompt}. Existing approaches using instruction templates typically use handcrafted or crowdsourced templates, resulting in a far smaller number of templates (hundreds to thousands), lower diversity, and fewer training instances compared to our work. 
\section{FineInstructions}

In this section, we detail the implementation of the FineInstructions pipeline illustrated in Figure \ref{fig:fipipeline}. Examples of generated outputs can be found in Appendix \ref{sec:appdx:illustratedexamples}. The FineInstructions pipeline takes as input a collection of user-written queries (questions, instructions, and or task requests). These queries are transformed into generic, reusable instruction templates that can be instantiated into a queries for many different subjects, topics, or domains. Given a set of pre-training documents, we match documents with compatible instruction templates. A document is considered a match if it contains enough information to both instantiate the query realistically as well as provide a grounded answer. Finally, we make use of a judge model to measure the quality of these instruction-answer pairs and filter down to a high-quality subset.

Our pipeline enables the creation of a large set of primarily human-written instruction-answer pairs that can be used to perform \textit{supervised} pre-training by conditioning answer generation on each instruction. This procedure closely resembles a weak supervision procedure where large, unlabeled corpora are transformed into large, supervised datasets with programmatic labeling \citep{ratner2017snorkel}. In this case, we use a large bank of instruction templates and existing models trained on some amount of supervised data to aid in extracting a larger quantity of supervised data. We use DataDreamer \citep{datadreamer}, a framework for synthetic data generation and training, to generate silver-standard data with task-specific prompts and train efficient distilled models that help perform each step of the pipeline at pre-training scale. An example of a task-specific prompt can be found in Appendix \ref{sec:appdx:promptexample}. Full details on our pipeline, prompts, training, and hyperparameters can be found in our code within the supplementary materials. The LLM we use to generate silver-standard data is Llama-3.3 70B Instruct \citep{llama3}. We discuss the implementation of each stage of this pipeline in greater detail in the following sections.

\subsection{Generating Instruction Templates}

We begin by collecting realistic instances of users querying LLMs, asking questions online (e.g., on forums), or searching via search engines. We also incorporate human-written prompt templates from popular prompt libraries. In total, we source \textasciitilde21.5M queries across a wide range of categories, including real-world LLM chat logs and search engine queries (e.g., \href{https://huggingface.co/datasets/allenai/WildChat}{WildChat}, \href{https://huggingface.co/datasets/lmsys/lmsys-chat-1m}{LMSys}, \href{https://huggingface.co/datasets/sentence-transformers/gooaq}{GooAQ}), community QA forums (e.g., \href{https://huggingface.co/datasets/ArmelR/stack-exchange-instruction}{StackExchange}, \href{https://huggingface.co/datasets/nreimers/reddit_question_best_answers}{Reddit QA}), domain-specific datasets (e.g., \href{https://huggingface.co/datasets/lavita/medical-qa-datasets}{ChatDoctor}), and standard instruction-tuning distributions (e.g., \href{https://huggingface.co/datasets/nguyenthanhdo/FLANv2-without-T0}{FLAN}, \href{https://huggingface.co/datasets/databricks/databricks-dolly-15k}{Dolly}). A comprehensive breakdown of all 20 data sources, including their sizes and references, is provided in Appendix \ref{sec:appdx:datasets}. After collection, we filter out harmful queries using the OpenAI Moderation API \citep{openai_moderation} and decontaminate from overlap with common benchmarks following Tulu 3 \citep{tulu3}, applying AI2's open-instruct scripts to index source queries and generated pairs in Elasticsearch and remove n-gram matches to benchmark test sets.

We convert queries to genericized templates by inserting \texttt{<fi></fi>} tags in place of spans in the query that refer to specific entities or scenarios. Within each pair of tags is a short, natural language description of what kind of text might be appropriate at that span. These spans are filled in with content from a matched document later in our pipeline. To train a model to convert queries into templates, we first select a subset of \textasciitilde 50K queries spanning the selected datasets. These queries are converted to ``silver'' synthetic templates by prompting an LLM with a series of prompts. We also generate synthetic ``compatible document descriptions'' for each of these \textasciitilde 50K templates. The description is a short natural language paragraph describing the kinds of documents that could instantiate such a template as well as provide an answer. We train the efficient Query Genericizer Model by fine-tuning a Llama-3.2 1B Instruct model on the generated synthetic data, which is then used to transform all \textasciitilde 18M queries into templates and compatible document descriptions. This pipeline ultimately produces \textasciitilde 18M templates.

\subsection{Matching Documents to Instruction Templates}

We next use a semantic similarity embedding model, BGE-M3, to embed the compatible document descriptions for the \textasciitilde 18M templates and build a FAISS retrieval index on the resulting embeddings \citep{bge_embedding,faiss}. Given a corpus of unstructured documents (e.g.\ a representative LLM pre-training dataset), we prompt an LLM to convert \textasciitilde 200K  documents into descriptions specifying the type of knowledge contained in each document, which are then embedded with the same model. With each document's embedding, we query the FAISS index and retrieve five potentially compatible instruction templates. We use the LLM to judge whether the document and instruction template are compatible to produce hard positive ($1$) and hard negative ($0$) examples of compatibility used with a cosine similarity loss to fine-tune the BGE-M3 embedding model. We fine-tune this model to directly embed documents and compatible document descriptions of instruction templates close together or far apart based on the compatibility labels \citep{sentence_transformers}. To further improve the task-relevance of the embeddings, we repeat this process, ultimately fine-tuning BGE-M3 twice. In the second fine-tuning stage, we introduce a custom pooling layer (described below) that allows us to efficiently produce multiple embeddings at the pooling layer when embedding a single document. Each of the $K$ pooled embeddings aims to represent each of $K$ different sections or chunks of the document. This allow us to retrieve templates relevant to different pieces of information throughout a long document.

\paragraph{Gaussian Pooling for Document Coverage} 
To ensure retrieved instruction templates adequately cover the entire document, we augment the standard global mean pooling with a custom ``Gaussian pooling'' layer. Instead of relying solely on a single global embedding, this layer computes multiple ``local'' embeddings using soft, Gaussian-weighted pooling. By evenly distributing $K = 5$ Gaussian kernels across the text sequence, each embedding captures a distinct semantic region (or ``chunk'') of the document, yielding 6 total retrieval representations per document (including the global embedding). 

In the second fine-tuning phase, we train the embedding model to reduce representation mixing and maintain strict local semantics. We achieve this by applying Gaussian weights to the input attention mask and training on hard positive and negative instruction templates explicitly labeled by their relevant document chunk index. This forces the pooling layer to reliably isolate local representations. Ultimately, this approach is highly effective: in our final FineInstructions dataset, we observe a Pearson correlation of 0.99 between the retrieval chunk index and the actual location of the answer excerpt within the source document (with most excerpts selected from the 19\%--71\% range, scaling linearly with the chunk index). We defer the rigorous mathematical formulation, pooling operations, and hyperparameter details to Appendix \ref{sec:appdx:gaussianpooling}.

\paragraph{Retrieval} When performing retrieval, we consider a template to be a match candidate for a given document if the cosine similarity between their embeddings is above a manually selected threshold ($0.865$). Among possible matches, we perform a weighted random sample to encourage diversity in the templates: Since our set of instruction templates contain both shorter, simple templates (with only one or two \texttt{<fi></fi>} tags) as well as longer, complex templates (with 10+ \texttt{<fi></fi>} tags), we compute sampling weights to match the distribution of the complexity of templates derived from real-world LLM queries in the \href{https://huggingface.co/datasets/allenai/WildChat}{WildChat}, \href{https://huggingface.co/datasets/lmsys/lmsys-chat-1m}{LMSys$\ast$}, \href{https://huggingface.co/datasets/OpenAssistant/oasst1}{OAsst1}, and \href{https://huggingface.co/datasets/shachardon/ShareLM}{ShareLM} datasets. 

\subsection{Generating Instructions and Answers}

Finally, with a sample of \textasciitilde 100K pre-training documents with six compatible instruction templates each, we prompt an  LLM to instantiate the instruction templates and identify excerpt(s) of the document that could be a relevant answer for the query. Since most documents are not written in the style of query responses, we allow the LLM to rephrase the excerpt slightly or add a few words at the beginning of the excerpt that more directly answers the query or instruction before providing more context and detail with an excerpt from the document. Since answers grounded in the document are more desirable than synthetically generated tokens \citep{recursivetraining}, we ensure that the ratio of excerpted text in generated answers is $\geq$0.80. Finally, we note that generating long responses in this stage is expensive since it involves token-by-token generation as well as possible GPU memory bottlenecks. We additionally note that 1) our answers are mainly direct excerpts from the document and 2) variables in the instruction template are mainly filled in by direct excerpts from the document. This allows us to reduce the computation required during synthetic data generation by introducing tags that designate that text should be directly copied instead of generated. For example, the text ``It is known that no preferred inertial frame exists according to the principle of relativity'' is generated simply as ``\texttt{<excerpt>}It is known that\texttt{<...>}the principle of relativity.\texttt{</excerpt>}''. In this way, the model only needs to generate \texttt{<excerpt>} tags with ellipsis notation (\texttt{<...>}) to reduce the number of generated tokens needed to excerpt long spans of text. These generated tags can be expanded inexpensively and programatically after decoding.

We create \textasciitilde 100K silver-standard examples and train a distilled model to instantiate instruction templates and generate answers by fine-tuning Llama-3.2 3B Instruct. With this efficient distilled model, we generate many instruction-answer pairs for a large number of documents and filter for high-quality instantiations and answers with judge prompts using the LLM until we are able to produce \textasciitilde 100K new high-quality instantiated instruction and answer examples. We keep some instances of when the LLM determines a retrieved instruction template is not compatible with the document as examples of when to output \texttt{null} instead of force a low-quality generation (\textasciitilde5\% of the total examples). We train the final Instantiator Model in a second round with these new examples. We ensure these \textasciitilde 100K new examples are also well stratified in length, complexity (number of template variables), and in topic using keyword-based filters to identify instruction-answer pairs related to math and code to ensure balanced representation in the distillation examples.

\subsection{Judging and Filtering Instructions and Answers}

Having pre-training data formatted as standalone instruction-answer pairs makes our generated data amenable to being used with off-the-shelf reward and judge models. We implement a judging and filtering stage on our synthetic instructions and answers to yield a higher-quality set. Specifically, we use the off-the-shelf Flow Judge model, a distilled 3.8B parameter judge model \citep{flowjudge} with a 5-point Likert scale rubric ranging from 1 (irrelevant or off-topic) to 5 (addresses the query without extraneous, vague, or repetitive content), and retain instruction-answer pairs that score $\geq 4$. In practice, the judge also filters malformed template instantiations and answers not grounded in the source document.
\section{Experimental Setup}

In this section, we describe our experimental setup to validate whether training from scratch on synthetic instruction and answer pairs from the FineInstructions pipeline improves knowledge absorption and model performance.

\subsection{Baselines}

For fair comparison, all methods we evaluate utilize the same unstructured document source corpora and all compared models are trained on the same number of tokens. As a first baseline, we consider training on the original documents themselves, as is done in standard pre-training pipelines. We also select a number of relevant baselines from prior work that propose similar procedures of converting pre-training documents with synthetic transformations or rephrasing. The authors of these prior works have released the vanilla pre-training corpora used in their experiments as well as their pre-computed synthetic transformations over the corpora. We describe them below.

\paragraph{Instruction Pre-Training (IPT)}
The ``Instruction Pre-Training'' method \citep{ipt} proposes training an instruction-and-response synthesizer model to convert documents into instruction-answer pairs. The synthesizer model is trained on instructions and responses from academic NLP Q\&A datasets. We use both the IPT vanilla pre-training corpus consisting of \textasciitilde23B tokens from RefinedWeb \citep{refinedweb} and the pre-computed synthetically transformed version of that corpus converted via the IPT instruction-and-response synthesizer model.

\paragraph{Nemotron-CC} The Nemotron-CC method \citep{nemotron} generates synthetic pre-training data by processing high-quality documents filtered from CommonCrawl \citep{commoncrawl} with an LLM using a mixture of tasks. These tasks include rephrasing documents, generating synthetic Q\&A pairs, and extracting, distilling, and listing core knowledge from the document. We use the Nemotron-CC vanilla pre-training corpus (\textasciitilde300B tokens) and the pre-computed Nemotron-CC synthetic pre-training data created using their mixture of tasks. We also select \textasciitilde300B tokens of pre-training data generated solely by their method of generating diverse Q\&A as a standalone baseline that is the most directly comparable to IPT and FineInstructions. Lastly, we select \textasciitilde300B tokens of their pre-computed synthetically rephrased pre-training data implementing the WRAP (Web Rephrase Augmented Pre-training) technique \citep{wrap} to compare with a strong rephrasing baseline.

\subsection{Pre-Training}

Using the original unstructured documents from both datasets, we use our pipeline to retrieve six instruction templates (with $K=5$ Gaussian pooling chunks to cover the document) and generate six instruction-answer pairs per document. For each pre-training document, we randomly keep instruction-answer pairs with a total token count that does not exceed the token count of the source document \footnote{If there is remaining token budget because instruction-answer pairs cannot exactly fill the token count of the source pre-training document, we rollover the remaining token budget to instruction-answer pairs for future pre-training documents.}. On average, this results in \textasciitilde 3 instruction-answer pairs per document. We format these instruction and answer pairs with a simple chat template similar to the other synthetic baselines: ``Instruction: \texttt{\{\{instruction\}\}}\textbackslash n\textbackslash nAnswer: \texttt{\{\{answer\}\}}'' (included in the token count). This allows us to produce a dataset-controlled, token-for-token, equivalent set of FineInstructions pre-training data that is \textasciitilde23B tokens and \textasciitilde300B tokens for the IPT and Nemotron-CC datasets respectively. Distinct from FineInstructions, the baseline methods do not produce standalone question-answer pairs and instead produce questions that only make sense when the source document is provided as context (e.g. ``What club does Helen like?''). They typically append their question-answer pairs at the end of the document (reading comprehension style Q\&A) and we pre-train them in their native format. We perform our pre-training experiments using the Lingua framework \citep{meta_lingua} designed for controlled pre-training ablations on 8xH100s and pre-train 1.8B parameter models with a Llama-3 tokenizer \citep{llama3} for each vanilla pre-training dataset, each baseline dataset, and our FineInstructions datasets. We train for a single epoch on datasets derived from Nemotron-CC and four epochs on datasets derived from IPT.

\subsection{Benchmarks}

We select three LLM evaluation benchmarks described below that aim to correlate with human judgments of model response quality. These benchmarks evaluate both knowledge absorption as well as the ability to respond to realistic user queries about recommendations, advice, suggestions, etc. When evaluating each method, we format the benchmark questions and instructions into a chat template that matches the method's training template and use the standard (``Instruction:'' and ``Answer:'') template for models pre-trained on vanilla pre-training data. We use greedy sampling when generating responses. Manually inspecting the responses for each method, we find all methods are able to respond and format answers to prompts reasonably well for accurate judging with few degenerate responses. 

\paragraph{MixEval} The MixEval benchmark \citep{mixeval} is a subset of task instances from a variety of academic LLM benchmarks such as TriviaQA \citep{mixeval_triviaqa}, MMLU \citep{mixeval_mmlu}, HellaSwag \citep{mixeval_hellaswag}, CommonsenseQA \citep{mixeval_commonsenseqa}, among others. MixEval aims to retain only examples from each of these datasets that best correlate with human judgments of response quality and grades responses with an LLM-as-judge given a reference answer. We use GPT-5 mini as a judge \citep{gpt5} and use the ``2024-08-11'' version on both the ``Standard'' and ``Hard'' splits.

\paragraph{MT-Bench-101} MT-Bench-101 is a multi-turn benchmark involving realistic, subjective user queries (opinions, advice, writing, etc.) graded via LLM-as-judge with a rubric scaled out of 10 \citep{mtbench101}. While it is a multi-turn benchmark, it is common to evaluate on this benchmark in single-turn fashion for models not tuned for multi-turn chat. We use GPT-5 mini as a judge \citep{gpt5}.

\paragraph{AlpacaEval} The AlpacaEval benchmark evaluates on realistic queries and tasks a user may ask a LLM and evaluates with LLM-as-judge in a head-to-head evaluation between the responses from two models to produce a win rate \citep{alpacaeval}. It also corrects for length-bias in judging \citep{alpacaeval_length_controlled}. We use the supported GPT-4-Turbo model as our judge model \citep{gpt4}.
\section{Results}

\fibenchmarktablescores

The results of our pre-training experiments can be found in Table \ref{table:fibenchmarktablescores}. Pre-training on data from the FineInstructions pipeline outperforms both standard pre-training and the other synthetic baselines on both datasets. For example, we observe a \textasciitilde 69\% relative improvement on MixEval compared to standard pre-training on the IPT dataset and a \textasciitilde39\% improvement on Nemotron-CC. A per-sub-benchmark breakdown of MixEval and MixEval-Hard performance is provided in Appendix~\ref{sec:appdx:mixevalsubbenchmarks}. On AlpacaEval, outputs from training on FineInstructions are consistently preferred over outputs from all other pipelines. MT-Bench-101 is the only reference-free evaluation (with no gold reference answer or head-to-head comparison) resulting in evaluation scores that yield lower spread and differentiation between models. Nevertheless, our FineInstructions data achieves a higher score on MT-Bench-101 than any other method. Notably, these improvements hold across both knowledge-focused (MixEval) and open-ended evaluation (MT-Bench-101 and AlpacaEval) benchmarks, suggesting that FineInstructions yields more consistent generalization across both kinds of tasks. Both IPT and Nemotron-CC demonstrated fairly small improvement (0-2\%) on MixEval despite generating narrow instruction data in a highly similar style targeting the evaluation benchmark format (multiple choice, reading comprehension Q\&A, etc.). On more realistic evaluations correlated to human judgments of response quality on open-ended real-world tasks such as those in AlpacaEval, we find these other techniques yield subpar results compared with FineInstructions. We hypothesize this is supported by the relative diversity of queries in FineInstructions compared to other pipelines' focus on generating instructions for a narrow set of academic benchmark tasks like multiple choice and Q\&A. Finally, in Appendix \ref{sec:appdx:modelsizeexperimetn}, we experiment with pre-training models at various sizes (300M, 1.8B, and 7B parameters) on FineInstructions and the strongest-performing baseline (Nemotron-CC) and find training on FineInstructions reaches higher performance levels with equivalent token and compute budget, making it an optimal technique for producing data for pre-training capable, small, and efficient language models.
\section{Discussion}

In this section, we provide an analysis on the diversity of the generated instructions, discuss an additional experiment on ablating the judging stage, and sketch out future directions.

\subsection{Diversity of Instructions}
We analyze our generated instructions to ensure there is sufficient diversity with no over-representation of certain instruction templates. It is desirable that long-tail instruction templates (e.g. ``Give me free apps that convert a file from \texttt{<fi>format \#1</fi>} to \texttt{<fi>format \#2</fi>}'') that are only compatible with highly specific documents are represented along with simpler instruction templates compatible with many documents (e.g. ``Give me a summary of \texttt{<fi>topic</fi>}''). When applied to the base Nemotron pre-training dataset, 4.3M unique instruction templates were used to instantiate \textasciitilde1.08B total instructions. We find no single instruction template comprises more than 0.09\% of the generated instructions and that the majority of instruction templates are used to instantiate fewer than 1,000 instructions. Consequently, our data is highly diverse in task and task formats. Overall, we find a power-fit relationship between the number of pre-training documents ($x$) and the number of unique instruction templates ($y$) utilized at least once when generating instructions from those documents from our set of \textasciitilde18M instruction templates  as approximately: $y= 16,891 * x ^ {0.24}$ with an $r^2=0.96$. Around \textasciitilde 50\% of instructions are instantiated from a template derived from GooAQ, with Reddit QA (\textasciitilde 27\%), LMSys Chat (\textasciitilde 9\%), and WildChat (\textasciitilde 6\%) following behind. All other sources represent $\leq1\%$ each.

% \verbdiversityfigure

We also analyze the type of the instructions generated.  We use Llama-3.3 70B Instruct with zero-shot prompts to classify instruction templates into various non-mutually exclusive domains (science, math, code, medicine, etc.) and provide the results in Table \ref{table:categorydiversity} in Appendix \ref{sec:appdx:categoricaldiversity}. Additionally, we define categories for measuring whether an instruction is a task that requires reasoning over knowledge (e.g. ``Compare and contrast \texttt{<fi>entity A</fi}> and \texttt{<fi>entity B</fi}>'') as well as if the instruction is ``tasky'' (e.g. ``Write a critique of \texttt{<fi>media performance</fi}>'') vs. a simple knowledge recall Q\&A task. Finally, we use the LLM to generate a short, one-sentence description of each instruction template describing the task at hand and visualize these sentences in a sunburst chart for a subset of instructions in Figure \ref{fig:verbdiversity} in Appendix \ref{sec:appdx:categoricaldiversity} following \citet{Köksal:24longform}. This kind of analysis with zero-shot classification may yield errors, but provides some high-level insights into the composition of tasks. The full annotation prompts can be found in our code.

\subsection{Effect of Judging and Filtering}

We conduct an ablation experiment to measure the effect of the  final judging and filtering stage. We pre-train with and without judging and filtering and provide these results in Appendix \ref{sec:appdx:judgingablation} on the IPT and Nemotron-CC datasets and find this added stage overall further improves the performance, especially on the AlpacaEval benchmark.

\subsection{Limitations and Future Directions}

We intentionally do not optimize for or report perplexity, a standard proxy for language modeling ability.
Because FineInstructions pre-trains models exclusively on instruction-answer pairs rather than raw next-token prediction over documents, we openly concede that standard pre-training likely outperforms our approach on perplexity-based metrics.
Our primary optimization target is downstream user utility (producing useful responses to prompts), rather than intermediate language modeling proxies, and our evaluation suite reflects this objective. Our comparisons against baselines may appear format-mismatched or that our approach has a supervision advantage. However, our approach and the synthetic baselines (IPT, Nemotron-CC, WRAP, and Q\&A) operate on the same principle: converting raw pre-training documents into a query-response training format. The performance gains we report over these baselines therefore stem from the design of our pipeline (diverse user-derived templates for more realistic queries, document-grounded excerpt answers, and judge filtering).

The FineInstructions pipeline could be optimized and scaled to yield better results. Several procedural choices, including 50K-example distillation sets and a retrieval threshold of 0.865, reflect compute budget constraints and manual calibration rather than systematic tuning. The distribution of source queries, calibration of the matching embedding, and the sampling weights all have influence on the composition and complexity of instructions generated. An optimal mixture may yield further improvements in performance. One common failure mode we observe in the generated instructions is that complex templates are challenging to match and instantiate. Scaling to larger models beyond the 1B and 3B scales considered in this work could yield stronger performance on complex templates and longer documents. Although our chosen benchmarks provide a reliable picture of language model performance in a range of scenarios, we also found that there is a paucity of benchmarks targeting the kind of long-tail realistic knowledge tasks users ask LLMs (recommendations, advice, suggestions, etc.) as opposed to factual recall of knowledge. Moreover, pre-training on instruction-answer pairs yields a model that consistently produces long-form answers and assigns a low probability to answer choices or short-form responses. This makes benchmarks using log probability-based classification for evaluation ill-suited for our setting and prior studies have found such evaluation is not reliable \citep{logproblimits1,logproblimits2}. Instead, extractive or LLM-as-judge based grading is recommended, which we follow. There are comparatively few LLM benchmarks in the latter category.
\section{Conclusion}

FineInstructions provides an effective pipeline for transforming real user queries into templates to generate in-distribution synthetic data at scale. We demonstrate that the resulting data can be effectively used to train LLMs in a supervised, instruction-aligned format rather than through self-supervised next token prediction on pre-training documents. By transforming the learning objective and the structure of pre-training data, this approach trains models on data that better reflects downstream usage patterns and improves knowledge absorption efficiency.
\section*{Acknowledgements}

We would like to acknowledge the Hugging Face organization and team for providing compute and storage resources for the experiments in this work and providing feedback on scaling the procedure. Particular thanks to Lewis Tunstall, Hynek Kydlíček, and Joel Niklaus for discussions around evaluation and related work. Thank you to our reviewers for their comments in strengthening this paper.

This research is based upon work supported in part by the Defense Advanced Research Projects Agency's (DARPA) SciFy program (Agreement No. HR00112520300) and by the Office of the Director of National Intelligence (ODNI), Intelligence Advanced Research Projects Activity (IARPA), via 56000026C0019. The views and conclusions contained herein are those of the authors and should not be interpreted as necessarily representing the official policies, either expressed or implied, of DARPA, ODNI, IARPA, the Department of Defense, or the U.S. Government. The U.S. Government is authorized to reproduce and distribute reprints for governmental purposes notwithstanding any copyright annotation therein.

\section*{Ethics Statement}
This paper presents work whose goal is to advance large language model training with synthetically generated data. One positive consequence of our work is that it allows for more efficient training with the synthetic corpus that only needs to be transformed and generated once to yield better results and faster convergence on subsequent training runs. Training on synthetically generated data may amplify biases and errors from the generating model. We mitigate this effect by taking near-exact excerpts from source documents and mainly using the generating model to transform naturally occurring text data into the desired format, rather than generate new content that could contain hallucinated content.  However, some marginal risk of systematic data biasing may still remain due to the scale of generation in this work.

\bibliography{other,custom}
\bibliographystyle{colm2026_conference}

\appendix
% Appendices
\appendix\onecolumn
\section{Dataset Sources}
\label{sec:appdx:datasets}
Below we detail the comprehensive list of datasets used to source user-written queries we convert into instruction templates.

\begin{table}[ht]
\centering
\small
\begin{tabular}{lrl}
\toprule
\textbf{Dataset} & \textbf{Query Count} & \textbf{Citation} \\
\midrule
\href{https://huggingface.co/datasets/ArmelR/stack-exchange-instruction}{StackExchange Instructions} & ~10.6M & -  \\
\href{https://huggingface.co/datasets/allenai/WildChat}{WildChat} & 657K & \citep{source_wildchat} \\
\href{https://huggingface.co/datasets/lmsys/lmsys-chat-1m}{LMSys Chat} & 559K & \citep{source_lmsys_chat} \\
\href{https://huggingface.co/datasets/lmsys/chatbot_arena_conversations}{LMSys Chatbot Arena Conversations} & 26.5K & \citep{source_lmsys_chatbot_arena_conversations} \\
\href{https://huggingface.co/datasets/OpenAssistant/oasst1}{OAsst1} & 3.92K & \citep{source_oasst1} \\
\href{https://huggingface.co/datasets/HuggingFaceH4/no_robots}{HuggingFace NoRobots} & 9.49K & \citep{source_HuggingFaceH4_no_robots} \\
\href{https://huggingface.co/datasets/nvidia/HelpSteer}{HelpSteer} & 10K & \citep{source_HelpSteer} \\
\href{https://huggingface.co/datasets/databricks/databricks-dolly-15k}{Dolly} & 14.7K & \citep{dolly} \\
\href{https://huggingface.co/datasets/nreimers/reddit_question_best_answers}{Reddit QA} & 7.47M & - \\
\href{https://huggingface.co/datasets/sentence-transformers/gooaq}{GooAQ} & 3.01M & \citep{source_gooaq} \\
\href{https://huggingface.co/datasets/shachardon/ShareLM}{ShareLM} & 264 & \citep{source_sharelm} \\
\href{https://huggingface.co/datasets/cmalaviya/expertqa}{ExpertQA} & 1.73K & \citep{source_expertqa} \\
\href{https://huggingface.co/datasets/lavita/medical-qa-datasets}{ChatDoctor iCliniq} & 7.32K & \citep{source_icliniq_healthcaremagic} \\
\href{https://huggingface.co/datasets/lavita/medical-qa-datasets}{ChatDoctor HealthcareMagic} & 112K & \citep{source_icliniq_healthcaremagic} \\
\href{https://huggingface.co/datasets/fka/awesome-chatgpt-prompts}{Awesome ChatGPT Prompts} & 203 & - \\
\href{https://docs.anthropic.com/en/prompt-library/}{Anthropic Prompt Library} & 64 & - \\
\href{https://github.com/langchain-ai/langchain}{LangChain Prompts} & 51 & \citep{source_langchain} \\
\href{https://huggingface.co/datasets/Muennighoff/natural-instructions}{NaturalInstructions} & 757 & \citep{naturalinstructions} \\
\href{https://huggingface.co/datasets/bigscience/P3}{P3} & 544 & \citep{t0} \\
\href{https://huggingface.co/datasets/nguyenthanhdo/FLANv2-without-T0}{FLAN} & 3.44K & \citep{flan} \\
\bottomrule
\end{tabular}
\caption{Overview of the data sources, their respective query counts, and citations used in our collection process.}
\label{tab:dataset_sources}
\end{table}
\newpage \section{Examples of FineInstructions}
\label{sec:appdx:illustratedexamples}

\fiillustratedexamplesfigure
\section{Gaussian Pooling Layer for Template $\Leftrightarrow $ Document Matching Embedding Model}
\label{sec:appdx:gaussianpooling}

Specifically, let $H = [h_1, \dots, h_T]$ denote the sequence of token embeddings with attention mask $m_t \in \{0,1\}$, where $T$ is the effective sequence length. We define $K$ as the number of Gaussians representing different chunks of the document, $c_k$ as the chunk center positions, $\rho_k = \frac{k}{K+1}$ as their normalized positions, $\sigma$ as the Gaussian width parameter, and $\alpha \in [0,1]$ as the blending weight. 

In our implementation, we choose $K = 5$ chunks, $\alpha = 1.0$, and $\sigma = 0.05$. Using $H$, $m_t$, and $T$, we compute a global mean embedding and $K$ Gaussian-weighted chunk embeddings centered at $c_k = \rho_k T$ with width $\sigma$. Each chunk embedding is optionally blended with the global embedding via $\alpha$. 

During the second round of fine-tuning, to train the model to maintain local semantic representations, we create training instances by applying Gaussian weights at the input attention mask. We only attend to tokens with a weight of $\geq 0.5$, producing $K$ distinct local embeddings with these attention masks to retrieve and compute the loss against hard negative and hard positive examples.

\begin{longtable}[h!]{p{13.5cm}}
\toprule
\textbf{Input Token Embeddings and Attention Mask:} \\[2pt]
$H = [h_1, \dots, h_T], \; h_t \in \mathbb{R}^d, \quad m_t \in \{0,1\}$ \\[6pt]

\textbf{Global Embedding:} \\[2pt]
$\bar{h} = \dfrac{\sum_{t=1}^{T} m_t h_t}{\sum_{t=1}^{T} m_t}$ \\[6pt]

\textbf{Gaussian Centers for $K$ Chunks:} \\[2pt]
$c_k = \rho_k T, \quad \rho_k = \dfrac{k}{K+1}, \; k = 1, \dots, K$ \\[6pt]

\textbf{Gaussian Weights:} \\[2pt]
$w_{k,t} = \dfrac{m_t \exp\!\left(-\dfrac{1}{2}\left(\dfrac{t - c_k}{\sigma T}\right)^2\right)}%
{\sum_{t'=1}^{T} m_{t'} \exp\!\left(-\dfrac{1}{2}\left(\dfrac{t' - c_k}{\sigma T}\right)^2\right)}$ \\[6pt]

\textbf{Chunk-Local Embeddings:} \\[2pt]
$\tilde{h}_k = \sum_{t=1}^{T} w_{k,t} h_t$ \\[6pt]

\textbf{Blend Chunk-Local Embeddings with Global Embedding:} \\[2pt]
$h_k^{*} = (1-\alpha)\bar{h} + \alpha \tilde{h}_k$ \\[6pt]

\textbf{Final Global and $K$ Chunk-Local Embeddings:} \\[2pt]
$E = [\,\bar{h},\, h_1^{*},\, \dots,\, h_K^{*}\,] \in \mathbb{R}^{d(K+1)}$ \\[2pt]
\bottomrule
\end{longtable}
\section{Example of Prompt in FineInstructions Pipeline}
\label{sec:appdx:promptexample}

\fipromptexamplefigure
\newpage
\section{Diversity of FineInstructions}
\label{sec:appdx:categoricaldiversity}

We visualize the diversity of instructions and tasks in FineInstructions below.

\verbdiversityfigure

 We note datasets like FineInstructions can be used to mine for large amounts of task-specific or domain-specific instructions. For example, with 1B+ instructions, even 0.58\% of the instructions being math instructions yields 6M+ math instructions. This could enable experiments on training with different mixtures of topics and task types as well as experiments with increasing the proportion of pre-training data on specific tasks to build a domain-specialist LLM which some prior work has found to be important to downstream performance \citep{futurework_singletask1,futurework_singletask2}.

\categorydiversitytable

\newpage

\section{Ablation of Judging and Filtering Stage}
\label{sec:appdx:judgingablation}

\fibenchmarktablejudgeablation
\section{Performance of Training on FineInstructions at Different Model Scales}
\label{sec:appdx:modelsizeexperimetn}

\modelsizeexperimentscores
\section{Detailed Breakdown of Sub-Benchmarks in MixEval}
\label{sec:appdx:mixevalsubbenchmarks}

Table~\ref{table:mixevalsubbenchmarks} and Table~\ref{table:mixevalhardsubbenchmarks} provide a per-sub-benchmark breakdown of MixEval and MixEval-Hard performance for all pre-training methods evaluated in this work. Sample counts ($n$) for each sub-benchmark are shown in the column headers.

\mixevalsubbenchmarkstable

\mixevalhardsubbenchmarkstable

\end{document}